\documentclass[final]{cvpr}

\usepackage{times}
\usepackage{epsfig}
\usepackage{graphicx}
\usepackage{amsmath}
\usepackage{amssymb}
\usepackage{float}
\usepackage{multirow}
\usepackage{subfigure}
\usepackage{color}
\usepackage{bbding}

\usepackage[pagebackref=true,breaklinks=true,colorlinks,bookmarks=false]{hyperref}

\def\cvprPaperID{3279} 
\def\confYear{CVPR 2021}

\begin{document}
\title{Intra-Inter Camera Similarity for Unsupervised Person Re-Identification}

\author{Shiyu Xuan \ \quad Shiliang Zhang\\
Department of Computer Science, School of EECS, Peking University\\
Beijing 100871, China\\
{\tt\small shiyu\_xuan@stu.pku.edu.cn, slzhang.jdl@pku.edu.cn}
}

\maketitle

\pagestyle{empty}  
\thispagestyle{empty} 
\begin{abstract}
\renewcommand{\thefootnote}{}

Most of unsupervised person Re-Identification (Re-ID) works produce pseudo-labels by measuring the feature similarity without considering the distribution discrepancy among cameras, leading to degraded accuracy in label computation across cameras. This paper targets to address this challenge by studying a novel intra-inter camera similarity for pseudo-label generation. We decompose the sample similarity computation into two stage, \emph{i.e.}, the intra-camera and inter-camera computations, respectively. The intra-camera computation directly leverages the CNN features for similarity computation within each camera. Pseudo-labels generated on different cameras train the re-id model in a multi-branch network. The second stage considers the classification scores of each sample on different cameras as a new feature vector. This new feature effectively alleviates the distribution discrepancy among cameras and generates more reliable pseudo-labels. We hence train our re-id model in two stages with intra-camera and inter-camera pseudo-labels, respectively. This simple intra-inter camera similarity produces surprisingly good performance on multiple datasets, \emph{e.g.}, achieves rank-1 accuracy of 89.5\% on the Market1501 dataset, outperforming the recent unsupervised works by 9+\%, and is comparable with the latest transfer learning works that leverage extra annotations. \footnote{The code is available at \url{https://github.com/SY-Xuan/IICS}.}
 	
\end{abstract}

	%

\section{Introduction} \label{introduction}
Person Re-Identification (ReID) aims to match a given query person in an image gallery collected from non-overlapping camera networks \cite{market1501, dukemtmc}.
Thanks to the powerful deep Convolutional Neural Network (CNN), great progresses have been made in fully-supervised person ReID \cite{zhang2017alignedreid, PCB, LuoSB, liu2019self, demo}. To relieve the requirement of expensive person ID annotation, increasing efforts are being made on unsupervised person ReID~\cite{zhai2020multiple, zou2020joint, jin2020global, yu2019unsupervised, wu2019unsupervised, fu2019self}, \emph{i.e.}, training with labeled source data and unlabeled target data, or fully relying on unlabeled target data for training.

\begin{figure}[!t]
	\centering
	\includegraphics[width=1\linewidth]{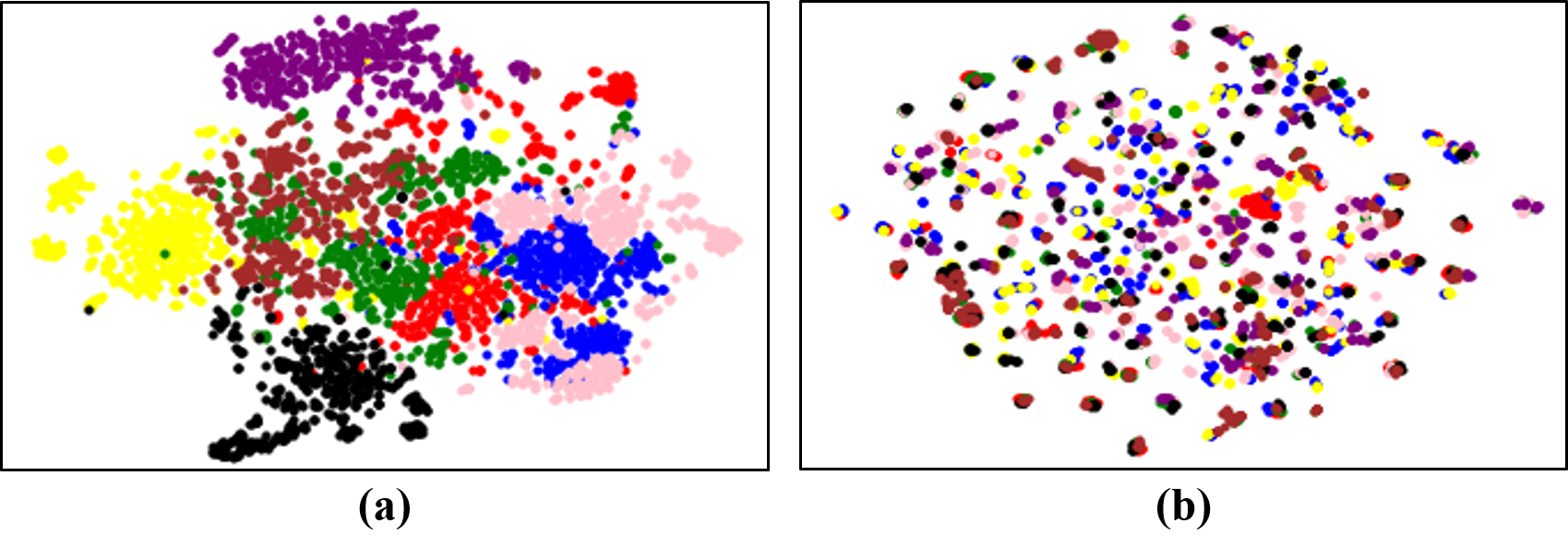}
	\caption{t-SNE visualization \cite{t-SNE} of features from a subset of DukeMTMC-ReID. Different colors indicate samples from different cameras. Baseline features in (a) suffer from feature distribution discrepancies among cameras. Features learned by our method are visualized in (b), where features from different cameras have similar distribution.}
	\label{fig:1}
\end{figure}
Existing unsupervised person ReID works can be grouped into three categories: a) using domain adaptation to align distributions of features between source and target domains~\cite{wu2019unsupervised, lin2018multi, wang2018transferable}, b) applying Generative Adversarial Network (GAN) to perform image style transfer, meanwhile maintaining the identity annotations on source domains~\cite{zou2020joint, PTGan, zhong2018camera, deng2018image}, and c) generating pseudo-labels on target domains for training via assigning similar images with similar labels via clustering, KNN search, \emph{etc.}~\cite{lin2019bottom, fan2018unsupervised, zhai2020ad, fu2019self, multi-label}. The first two categories define unsupervised person ReID as a transfer learning task, which leverages the labeled data on source domains. Generating pseudo-labels makes it possible to train ReID models with fully unsupervised setting, thus shows better flexibility.

Most of pseudo-labels prediction algorithms share a similar intuition, \emph{i.e.}, first computing sample similarities, then assigning similar samples identified by clustering or KNN with similar labels. During this procedure, the computed sample similarity largely decides the ReID accuracy. To generate high quality pseudo-labels, samples of the same identity are expected share larger similarities than with those from different identities. However, the setting of unsupervised person ReID makes it difficult to learn reliable sample similarities, especially for samples from different cameras. For example, each identity can be recorded by multi-cameras with varied parameters and environments. Those factors may significantly change the appearance of the identity. In other words, the domain gap among cameras makes it difficult to identify samples of the same identity, as well as to optimize of intra-class feature similarity. We illustrated the feature distribution of different cameras in Fig.~\ref{fig:1} (a).

This paper addresses the above challenge by studying a more reasonable similarity computation for pseudo-labels generation. Identifying samples of the same identify within the same camera is easier than performing the same task among different cameras. Meanwhile, domain gaps can be alleviated by learning generalizable classifiers. We hence decompose the sample similarity computation into two stages to progressively seek reliable pseudo-labels. The first stage computes sample similarity within each camera with CNN features. This ``intra-camera'' distance guides pseudo-label generation within each camera by clustering samples and assigning samples within the same cluster with the same label. Independent pseudo-labels in $C$ cameras hence train the ReID model with a $C$-branch network, where the shared backbone is optimized by multiple tasks, and each branch is optimized by a specific classification task within the same camera. This stage simplifies pseudo-label generation, thus ensures high quality pseudo-labels and efficient backbone optimization.

The second stage proceeds to compute sample similarities across cameras. Sample similarity computed with CNN features can be affected by domain gap, \emph{e.g.}, large domain gap decreases the similarity among samples of the same identity as illustrated in Fig.~\ref{fig:1} (a). As discussed in previous works~\cite{dou2019domain, tzeng2015simultaneous}, the classification probability is more robust the domain gap than raw features. We alleviate the domain gap by enhancing the generalization ability of trained classifiers in the first stage. Specifically, we classify each sample with $C$ classifiers, and use their classification scores as a new feature vector. To ensure the classification scores robust to the domain gap, each classifier trained on one camera should generalize well on other cameras. This is achieved with the proposed Adaptive Instance and Batch Normalization (AIBN), which enhances the generalization ability of classifier without reducing their discriminative ability. Classification scores produced by ${C}$ classifiers are hence adopted to calculate the ``inter-camera'' similarity to seek pseudo-labels across cameras. The ReID model is finally optimized by pseudo-labels generated with both stages. Distribution of features learned by our method is illustrated in Fig.~\ref{fig:1} (b), where the domain gaps between cameras are effectively eliminated.

We test our approach in extensive experiments on multiple ReID datasets including Market1501~\cite{market1501}, DukeMTMC-ReID~\cite{dukemtmc} and MSMT17~\cite{PTGan}, respectively. Experiments show that each component in our approach is valid in boosting the ReID performance. A complete approach consisting of intra-inter camera similarities exhibits the best performance. For instance, without leveraging any annotations, our approach achieves rank-1 accuracy of 89.5\% on the Market1501 dataset, outperforming the recent unsupervised works by 9+\%. Our method also performs better than many recent transfer learning works that leverage extra annotations. For instance, the recent MMT~\cite{MMT} and NRMT~\cite{zhao2020unsupervised} achieves lower rank-1 accuracies of 87.7\% and 87.8\% respectively, even they leverage extra annotations on DukeMTMC-ReID~\cite{dukemtmc} for training.

The promising performance demonstrates the validity of our method, which decomposes the similarity computation into two stages to progressively seek better pseudo-labels for training. This strategy is more reasonable than directly predicting pseudo-labels across cameras in that, it effectively alleviates the domain gap between cameras. Besides that, those two stages corresponds to different difficulty in predicting pseudo-labels, thus are complementary to each other in optimizing the ReID model. To the best of our knowledge, this is an original work studying better similarity computation strategies in unsupervised person ReID.

\section{Related Work}


This work is closely related to unsupervised person ReID and, domain adaptation and generalization. Recent works on those two topics will be reviewed briefly in following paragraphs.

\textbf{Unsupervised person ReID} has been studied with three types of methods, \emph{i.e.}, by distribution alignment, training GANs, and generating pseudo-labels, respectively. Distribution alignment based methods follow the traditional domain adaptation methods~\cite{MMD, coral} to align the feature distribution of source and target domains. Wu \etal~\cite{wu2019unsupervised} proposed a Camera-Aware Similarity Consistency Loss to align the pairwise distribution of intra-camera matching and cross-camera matching.
Lin \etal~\cite{lin2018multi} utilized Maximum Mean Discrepancy (MMD) distance \cite{MMD} to align the distribution of mid-level features from source and the target domains. Some other methods use GANs~\cite{CycleGAN2017} to perform image-to-image style translation to transfer source images into target style.
Zou \etal~\cite{zou2020joint} disentangled id-related/unrelated features to enforce the adaptation to work on the id-related feature space.
Wei \etal~\cite{PTGan} proposed person transfer GAN, which can transfer person images with the style of target dataset and keep the identity label of the person.

Pseudo-labels based methods first generate pseudo-labels by formulating certain rules based on sample similarity, then train the ReID model with those pseudo-labels.
The quality of computed pseudo-labels determines the performance of these methods. Unsupervised clustering method is one of the most commonly used methods to generate pseudo-labels \cite{jin2020global, zhang2019self, multi-label, zhai2020ad, liu2020domain}.
Fan \etal~\cite{fan2018unsupervised} used standard k-means clustering method to generate pseudo-labels and fine-tuned model with these labels.
Lin \etal~\cite{lin2019bottom} proposed a bottom-up clustering approach to generate pseudo-labels. To avoid re-initializing the classifier at each epoch, an extra memory bank was added into the network.
Wang \etal~\cite{multi-label} formulated unsupervised person ReID as multi-label classification task and used memory bank to train the network.
NRMT \cite{zhao2020unsupervised}, MMT \cite{MMT} and MEB-Net \cite{zhai2020multiple} used mutual-training \cite{mutual} to reduce the influence of low-quality pseudo-labels.


\textbf{Domain adaptation and generalization} are commonly considered to improve the generalization ability of CNN models. Recently, some works have found that Batch Normalization (BN) \cite{BN} and Instance Normalization (IN) \cite{IN} could improve the network's generalization ability on multiple domains~\cite{zhuangrethinking, IBNNet, chang2019domain}. 
IBN-Net \cite{IBNNet} integrated IN and BN to enhance the generalization capacity of CNNs to unseen domain without fine-tuning. Chang \etal~\cite{chang2019domain} improved the performance of unsupervised domain adaptation using domain-specific BN. Zhuang \etal~\cite{zhuangrethinking} designed a camera-based BN to alleviate the distribution gap between a camera pair in person ReID. Their method improved the generalization ability of the model across unseen cameras. 

Most pseudo-labels based methods try to mitigate the impact of low-quality pseudo-labels or find high-quality part from generated pseudo-labels.
The work most similar to us is \cite{zhu2021intracamera} which utilizes extra ID labels within each camera as supervision and simply uses classification results to find matching candidates across cameras during inter-camera training.
Different from those works, our work is motivated to seek a reliable similarity by progressively eliminating negative influences of pose variances, illumination, occlusions through intra-camera training, and domain gap through inter-camera
training. This leads to the proposed AIBN and inter-camera similarities.
As shown in our experiments, our method produces surprisingly good performance on multiple datasets.


\section{Methodology}
\subsection{Formulation} \label{sec:formulate}

\begin{figure}
	\centering
	\vspace{-1mm}
	\includegraphics[width=0.95\linewidth]{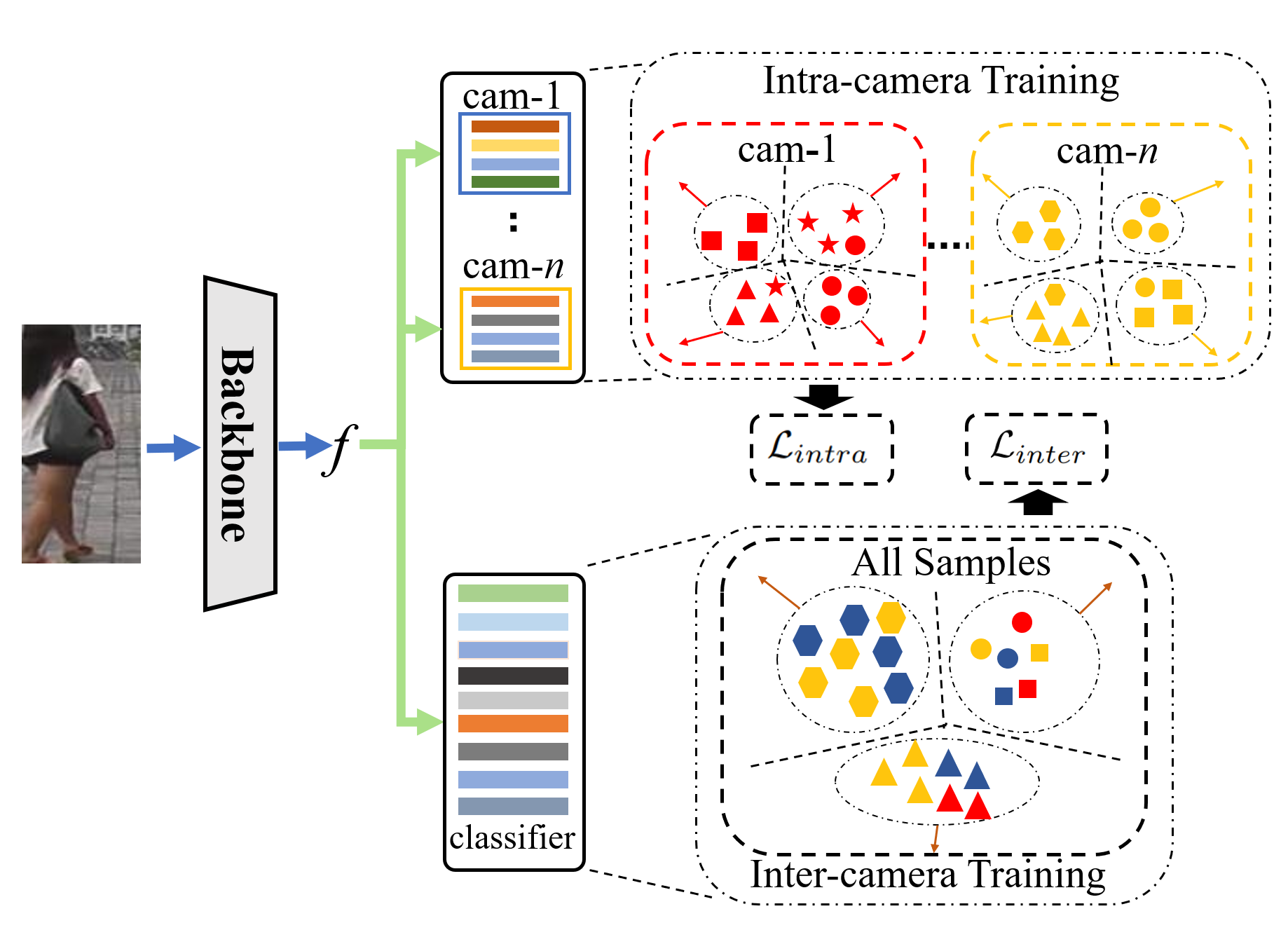}
	\caption{Illustrations of the proposed method for unsupervised person ReID. The Intra-camera training is conducted within each camera separately. It generates pseudo-labels by clustering using intra-camera similarity computed with the CNN feature $\boldsymbol{f}$. The inter-camera training generates pseudo-labels by clustering all samples using the inter-camera similarity, which is computed with classification scores. These two stages are executed alternately during the whole training process to optimize the ReID feature $\boldsymbol{f}$ with complementary intra and inter camera losses.}
	\label{fig:detail_pipeline}
\end{figure}

Given an unlabeled person image dataset with camera information $\mathcal X=\{\mathcal X^c\}$, where $\mathcal X^c$ is a collection of person images and the superscript ${c=1:C}$ denotes the index of cameras, respectively. Our goal is to train a person ReID model on $\mathcal X$. For any query person image $q$, the ReID model is expected to produce a feature vector to retrieve image $I_g$ containing the same person from a gallery set $G$. The trained ReID model should guarantee $q$ share more similar feature with $I_g$ than with other images in $G$, \emph{i.e.},
\begin{equation} \label{eq:reid}
g^* = \arg\max_{g\in G} \operatorname {sim}(\boldsymbol f_g,\boldsymbol f_q),
\end{equation}
where $\boldsymbol{f} \in \mathbb{R}^d$ is a $d$-dimensional feature vector extracted by the person ReID model. $\operatorname {sim} (\cdot)$ computes the feature similarity.

Suppose a person $p$ is captured by cameras in $\mathcal X$, the collection of images of $p$ and $\mathcal X$ can be denoted as $\mathcal X_p$ and $\mathcal X = \{\mathcal X_p\}_{p=1:P}$, respectively, where $P$ is the total number of persons in $\mathcal X$. An estimation towards $\{\mathcal X_p\}_{p=1:P}$ would make the optimization to Eq.~\eqref{eq:reid} possible, \emph{e.g.}, through minimizing feature distance within each $\{\mathcal X_p\}$, meanwhile enlarging distance between $\{\mathcal X_i\}$ and $\{\mathcal X_j\}$ with $i \neq j$. A commonly used strategy is performing clustering on $\mathcal X$ to generate pseudo-labels. The training objective in label prediction could be conceptually denoted as,
\begin{equation} \label{eq:clusterobj}
\mathcal T^* = \arg\min_{\mathcal T} \operatorname{\mathcal{D}}(\mathcal T, \{\mathcal X_p\}_{p=1:P}),
\end{equation}
where $\mathcal T$ denotes the clustering result and $\operatorname {\mathcal{D}}(\cdot)$ computes its differences with $\{\mathcal X_p\}_{p=1:P}$.

The optimization towards Eq.~\eqref{eq:clusterobj} requires to identify images of the same person across cameras. This could be challenging because the appearance of an image can be affected by complicated factors. Using $I_n^c \in \mathcal X^c$ to denote an image of person $p$ captured by camera $c$, we conceptually describe the appearance of $I_n^c$ as,
\begin{equation}
I_n^{c} \doteq A_p + S_c + E_n,
\label{eq:appearance}
\end{equation}
where $A_p$ denotes the appearance of the person $p$. $S_c$ represents the setting of cameras $c$ including its parameters, viewpoint, environment, \emph{etc.}, that affect the appearance of its captured images. We use $E_n$ to represent other stochastic factors affecting the appearance of $I_n^{c}$ including pose, illumination, occlusions, \emph{etc}.

According to Eq.~\eqref{eq:appearance}, the challenge of Eq.~\eqref{eq:clusterobj} lies in learning feature $\boldsymbol{f}$ to alleviate the effects of $S_c$ and $E_n$, and finding image clusters across cameras according to $A_p$. To conquer this challenge, we propose to perform pseudo-label prediction with two stages to progressively enhance the robustness of $\boldsymbol{f}$ to $E_n$ and $S_c$, respectively.

The robustness to $E_n$ can be enhanced by performing Eq.~\eqref{eq:clusterobj} within each camera using existing pseudo-label generation methods, then training $\boldsymbol{f}$ according to the clustering result. Suppose the clustering result for the $c$-th camera is $\mathcal T^c$, the training loss on $c$-th cameras can be represented as,
\begin{equation}
\mathcal L_{intra}^c = \sum_{I_n\in \mathcal X^c, I_n \in \mathcal T^c_m} \operatorname {loss}^c(\boldsymbol{f}_n, m),
\label{eq:intraloss}
\end{equation}
where $m$ denotes the cluster ID, which is used as the pseudo-label of $I_n$ for loss computation. To ensure the robustness of $\boldsymbol{f}$ towards complicated $E_n$ under different cameras, Eq.~\eqref{eq:intraloss} can be computed on different cameras by sharing the same $\boldsymbol{f}$. This leads to a multi-branch CNN, where each branch corresponds to a classifier, and their shared backbone learns the feature $\boldsymbol{f}$.

The robustness to $S_c$ is enhanced in the second stage by clustering images of the same person across cameras. Directly using the learned $\boldsymbol{f}$ to measure similarity for clustering suffers from $S_c$. We propose to compute a more robust inter-camera similarity. The intuition is to train classifiers with domain adaption strategies to gain enhanced generalization ability, \emph{e.g.}, the classifier on camera $c$ is expected to be discriminative on other cameras. We thus could identify images of the same person from different cameras based on their classification scores, and enlarge their similarity with the inter-camera similarity, \emph{i.e.},
\begin{equation}
\operatorname {SIM}_{inter} (I_m,I_n) = \operatorname {sim}(\boldsymbol{f}_m,\boldsymbol{f}_n) + \mu\Delta(\textbf{s}_m,\textbf{s}_n),
\label{eq:intersim}
\end{equation}
where $\textbf{s}_n$ denotes the classification score of image $I_n$. $\Delta(\textbf{s}_m,\textbf{s}_n)$ is the probability that $I_m$ and $I_n$ are from the same identity. Eq.~\eqref{eq:intersim} enlarges the similarity of two images from different cameras, if they are identified as the same person. It effectively alleviates $S_c$ during similarity computation and image clustering. We hence further optimize $\boldsymbol{f}$ with the inter-camera loss based on the clustering result $\mathcal T$, \emph{i.e.},
\begin{equation}
\mathcal L_{inter} = \sum_{I_n \in \mathcal T_m} \operatorname {loss}(\boldsymbol{f}_n, m).
\label{eq:interloss}
\end{equation}

Our method is progressively optimized by Eq.~\eqref{eq:intraloss} and Eq.~\eqref{eq:interloss}, respectively to gain $\boldsymbol{f}$ with the robustness to $S_c$, $E_n$. Their detailed computations, as well the implementation of $\Delta(\cdot)$ and generalization ability enhancement will be presented in the following parts.

\subsection{The Intra-camera Training}

Fig.~\ref{fig:detail_pipeline} illustrates our framework, where the person ReID feature $\boldsymbol{f}$ is optimized by two stages. The intra-camera training stage divides the training set $\mathcal X$ into subsets $\{\mathcal X^c\}$ according to the camera index of each image. Then, it performs clustering on each subset according to the similarity computed with feature $\boldsymbol{f}$. Assigning images within each cluster with identical label turns each $\mathcal X^c$ into a labeled dataset, allowing the function $\operatorname {loss}^c(\cdot)$ in $\mathcal L_{intra}^c$ can be computed as
\begin{equation}
\operatorname {loss}^c(\boldsymbol{f}_n, m) = \ell\left(\mathcal{F}\left(\boldsymbol{w^{c}},\boldsymbol{f}_n \right),m\right),
	\label{intra_loss}
\end{equation}
where $\mathcal{F}(\boldsymbol{w^{c}}, \cdot)$ denotes a classier with learnable parameters $\boldsymbol{w^{c}}$. $\ell(\cdot)$ computes the softmax cross entropy loss on classifier outputs and the groundtruth label $m$.

As illustrated in Fig.~\ref{fig:detail_pipeline}, the intra-camera training treats each cameras as a training task and trains the $\boldsymbol{f}$ with multiple tasks. The overall training loss can be denoted as
\begin{equation} \label{eq:intra_loss_sum}
\mathcal L_{intra} = \sum_{c=1}^{{C}}\mathcal L_{intra}^c,
\end{equation}
where $C$ is the total number of cameras. As discussed in Sec.~\ref{sec:formulate}, Eq.~\eqref{eq:intra_loss_sum} effectively boosts the discriminative power of $\boldsymbol{f}$ within each camera. Besides that, optimizing $\boldsymbol{f}$ on multi-tasks boosts its discriminative power on different domains, which in-turn enhances the generalization ability of learned classifiers.

\subsection{The Inter-camera Training}\label{reduction}

To estimate the probability that two samples from different cameras belong to the same identity, a domain-independent feature is needed.
As discussed in related works~\cite{dou2019domain, tzeng2015simultaneous}, samples belonging to the same identity should have similar distribution of classification probability produced by each classifier. We use the jaccard similarity of classification probability to compute the $\Delta(\textbf{s}_m,\textbf{s}_n)$, which reflects the probability that $I_m$ and $I_n$ are from the same identity
\begin{equation}
	\Delta(\textbf{s}_m,\textbf{s}_n)= \frac{\textbf{s}_m \cap \textbf{s}_n}{\textbf{s}_m \cup \textbf{s}_n},
	\label{jaccard}
\end{equation}
where $\cap$ is the element-wise min of two vectors and $\cup$ is the element-wise max of two vectors. The classification score $\boldsymbol{s}_m$ is acquired by concatenating the classification scores from \emph{C} classifiers,
\begin{equation}
	\begin{split}	
 \boldsymbol{s}_m =& [\boldsymbol s_m^1, \cdots, \boldsymbol s_m^c], \\
	\textbf{s}_m^c=& [p(1|\boldsymbol{f}_{m}, \boldsymbol{w_{c}}), \cdots, p(k|\boldsymbol{f}_{m}, \boldsymbol{w_{c}})],
	\end{split}
	\label{sm}
\end{equation}
where $p(k|\boldsymbol{f}_{m}, \boldsymbol{w_{c}})$ is the classification probability of at class $k$ computed by the classifier $\mathcal{F}\left(\boldsymbol{w_{c}}; \cdot\right)$ and $\textbf{s}_m^c$ is the classification score of image $I_m$ on camera $c$.

To make the $\Delta(\textbf{s}_m,\textbf{s}_n)$ work as expected, classifier trained on each camera needs to generalize well on other cameras. The $\boldsymbol{f}$ trained by multi-task learning in the intra-camera stage provides basic guarantee for generalization ability of the network. In order to further improve generalization of different classifiers, we propose AIBN which will be described in detail at Sec. \ref{AIBN}.



With $\Delta(\boldsymbol{s}_m,\boldsymbol{s}_n)$, clustering can be performed based on inter-camera similarity to generate pseudo-labels on $\mathcal X$. Then Eq. \eqref{eq:interloss} can be computed as:
\begin{equation}
	\mathcal L_{inter} = \frac{1}{|\mathcal B|} \sum_{I_n\in \mathcal B} \ell\left(\mathcal{F}\left(\boldsymbol{w}, \boldsymbol{f}_n\right), m\right) + \lambda L_{triplet},
	\label{inter_camera_loss}
\end{equation}
where $\mathcal B$ is a training mini-batch, $\ell$ is the softmax cross entropy loss, $m$ is its pseudo-label assigned by clustering result, $\lambda$ is loss weight and $L_{triplet}$ is the hard-batch triplet loss~\cite{triplet}. We randomly select $P$ clusters and $K$ samples from each cluster to construct the training mini-batch $\mathcal B$.

\subsection{Adaptive Instance and Batch Normalization}\label{AIBN}

As discussed above, we propose AIBN to boost the generalization ability of learned classifiers. Instance Normalization (IN) \cite{IN} can make the network invariant to appearance changes. However, IN reduces the inter-class variance, making the network less discriminative. Different from IN, Batch Normalization (BN) \cite{BN} retains variations across different classes and reduces the internal covariate shift during network training. In other words, IN and BN are complementary to each other.

In order to gain the advantages of both IN and BN, we propose the AIBN. It is computed by linearly fusing the statistics (mean and var) obtained by IN and BN, \emph{i.e.},
\begin{equation}
	\hat{\mathbf{x}}[i, j, n]=\gamma \frac{\mathbf{x}[i, j, n]-(\alpha\mu_{bn}+(1-\alpha)\mu_{in})}{\sqrt{\alpha\sigma_{bn}^{2}+(1-\alpha)\sigma_{in}^{2}+\epsilon}} + \beta,
	\label{combine}
\end{equation}
where $\mathbf{x}[i, j, n] \in \mathbb{R}^{H \times W \times N}$ is the feature map of each channel, $\mu_{bn}$ and $\sigma_{bn}$ are the mean and variance calculated by BN, $\mu_{in}$ and $\sigma{in}$ are the mean and variance calculated by IN, $\gamma$ and $\beta$ are affine parameters and $\alpha$ is a learnable weighting parameter. The optimization of $\alpha$ can be conducted with back-propagation during CNN training. We add no constraints to $\alpha$ during training back-propagation. During network forward inference using Eq. \eqref{combine}, we clamp $\alpha$ into $[0, 1]$ to avoid negative values.

\begin{figure}
	\centering
	\includegraphics[width=0.95\linewidth]{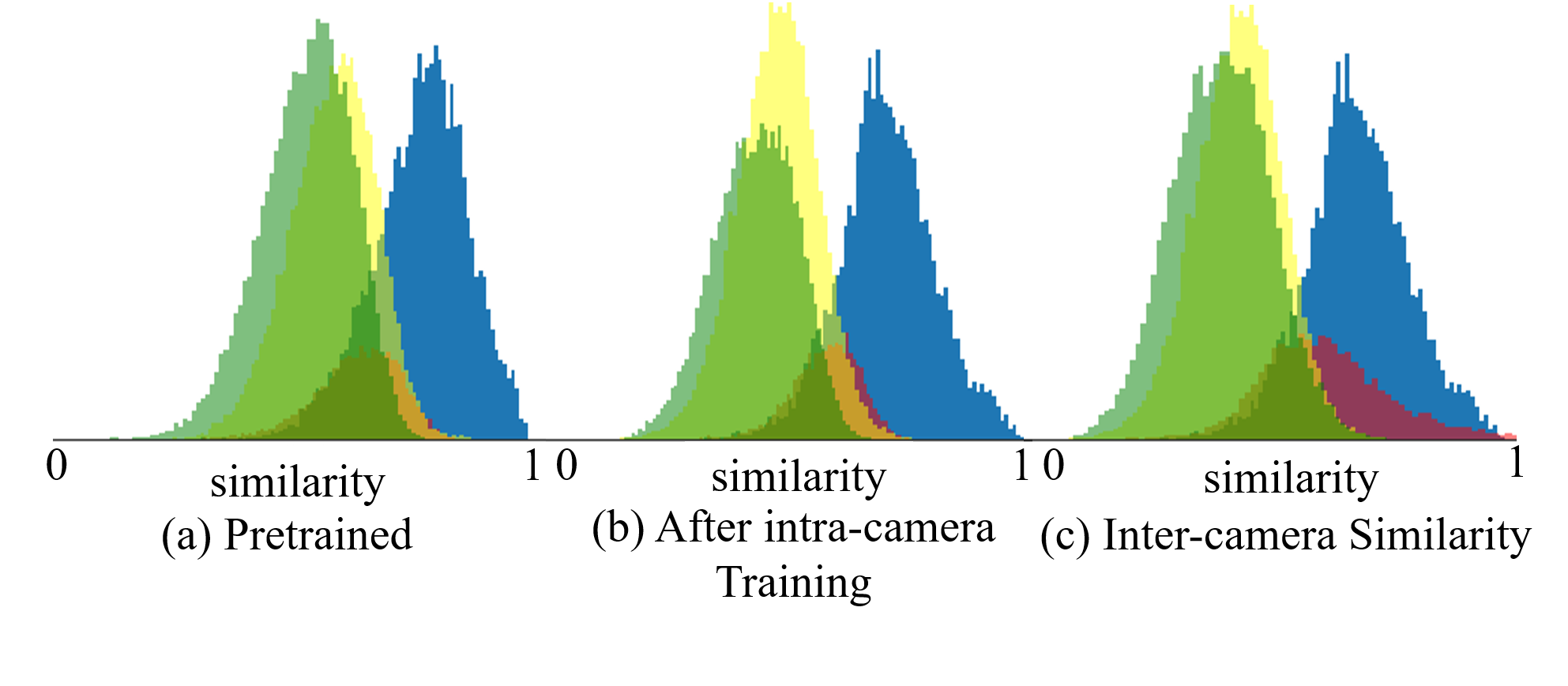}
	\vspace{-4mm}
	\caption{The distribution of similarity on DukeMTMC-ReID. \textcolor{blue}{Blue} and \textcolor{red}{Red} color indicates the distribution of similarity between samples of the same identity from the same camera and different cameras, respectively. \textcolor{yellow}{Yellow} and \textcolor{green}{Green} color indicates the distribution of similarity between samples of different identities from the same camera and different cameras, respectively. To show an intuitive visualization of real data, similarities are normalized into [0, 1].}
	\label{distribute}
\end{figure}

\textbf{Discussion:} To show the effects of two training stages, we visualize the distribution of similarities between samples in Fig. \ref{distribute}. We use \textcolor{red}{Red} color to indicate the distribution of similarity between samples from different cameras. It is clear in Fig. \ref{distribute} (a) that, the \textcolor{red}{Red} color is mixed with the \textcolor{yellow}{Yellow} and \textcolor{green}{Green} color, which indicate the distribution of similarity between samples of different identities. Therefore, clustering using similarity in Fig. \ref{distribute} (a) would lead to poor performance. It also can be observed that, the intra-camera training and inter-camera training progressively improves the discriminative power of feature similarity. The inter-camera training produces the most reliable similarity. More evaluations will be presented in the following section.

\section{Experiments}
\subsection{Dataset and Evaluation Metrics}
We evaluate our methods on three commonly used person ReID datasets, \eg, DukeMTMC-ReID \cite{dukemtmc}, Market1501 \cite{market1501}, and MSMT17 \cite{PTGan}, respectively.

\textbf{DukeMTMC-ReID} is collected from 8 non-overlapping camera views, containing 16,522 images of 702 identities for training, 2,228 images of the other 702 identities for query, and 17,661 gallery images.

\textbf{Market1501} is a large-scale dataset captured from 6 cameras, containing 32,668 images with 1,501 identities. It is divided into 12,936 images of 751 identities for training and 19,732 images of 750 identities for testing.

\textbf{MSMT17} is a newly published person ReID dataset. It contains 126,441 images of 4,101 identities captured from 15 cameras. It is divided into 32,621 images of 1,041 identities for training and 93,820 images of 3,060 identities for testing.

During training, we only use images and camera labels from the training set of each dataset and do not utilize any other annotation information. Performance is evaluated by the Cumulative Matching Characteristic (CMC) and mean Average Precision (mAP).

\subsection{Implementation Details}
We use ResNet-50 \cite{resnet} pre-trained on ImageNet \cite{imagenet} as backbone to extract the feature. The layers after pooling-5 layer are removed and a BN-Neck \cite{LuoSB} is added behind it. During testing and clustering, we extract the pooling-5 feature to calculate the similarity.
All models are trained with PyTorch.

During training, the input image is resized to $256 \times 128$. Image augmentation strategies such as random flipping and random erasing are performed. At each round we perform intra-camera stage and inter-camera stage in order. The number of training round is set as 40.

At intra-camera training stage, the batch size is 8 for each camera. The SGD is used to optimize the model. The learning rate for ResNet-50 base layers is $0.0005$, and the one for other layers is $0.005$.

At inter-camera training stage, a mini-batch of 64 is sampled with $P=16$ randomly selected clusters and $K=4$ randomly sampled images per cluster. The SGD is also used to optimize the model. The learning rate for ResNet-50 base layers is $0.001$, and the one for other layers is $0.01$. The loss weight $\lambda$ in Eq. \eqref{inter_camera_loss} is fixed to 1. Margin in triplet loss is fixed to 0.3. The training progressively uniforms the distribution of features from different cameras. Therefore, the initial $\mu$ in Eq. \eqref{eq:intersim} is set as 0.02, and follows the poly policy for decay.

For Market1501 and DukeMTMC-ReID, we train the model for 2 epochs at both stages. For MSMT17 we train the model for 12 epochs at intra-camera stage and 2 epochs at inter-camera stage.
We use the standard Agglomerative Hierarchical method \cite{scikit-learn} for clustering. For Market1501 and DukeMTMC-ReID, the number of clusters is 600 for each camera at intra-camera stage and 800 at inter-camera stage. For MSMT17, the number of clusters is 600 for each camera at intra-camera stage and 1200 at inter-camera stage.

Although additional clustering within each camera is performed, this is more efficient than clustering on the entire set. Therefore, the computational complexity of our method is acceptable. It takes about 4-5 hours to finish the training with a GPU on Market1501. 


For the AIBN, the mixture weight $\alpha$ is initialized to 0.5. We replace all BNs in layer3 and layer4 of ResNet-50 with AIBN. Mixture weights are shared at each BottleNeck module. The detailed analysis of this component is performed in Section. \ref{ablation}.
\subsection{Ablation Study}\label{ablation}
\paragraph{The impact of individual components.} In this section we evaluate the effectiveness of each component in our method, the experimental results of each setting are summarized in Table \ref{tab:components}.
As shown in the table, when only inter-camera stage is used for training, performance is not satisfactory. This shows that, the similarity between samples from different cameras is unreliable. Clustering directly using this similarity can lead to a poor performance.
The rank-1 accuracy on Market1501 and DukeMTMC-ReID can achieve 71.6\% and 62.9\%, respectively, with only intra-camera training stage. This indicates that the similarity between samples from the same camera is reliable.
Without considering the distribution gap between cameras, the addition of the inter-camera training stage leads to a decrease in performance on DukeMTMC-ReID. It is clear that although the feature produced by the model has been improved after the intra-camera stage, the similarity between samples from different cameras is still unreliable.

Our method achieves the best performance when the inter camera similarity in Eq. \eqref{eq:intersim} is used in inter-camera training stage. It demonstrates that the inter-camera similarity is more effective than CNN features similarity and is crucial for the our performance enhancement. 
Since the jaccard similarity can be used to calculate the probability that samples belong to the same identity, Eq. \eqref{jaccard} can also be used as similarity for inter-camera clustering. This setting can also achieve good performance, which means the jaccard similarity is also more robust to domain gap between cameras.
Experimental results show that each component in our method is important for performance boost, and their combination achieves the best performance.

\begin{table}
	\begin{center}
		\scalebox{0.9}{
		\begin{tabular}{c|cc|cc}
			\hline
			Dataset       & \multicolumn{2}{c|}{Market } & \multicolumn{2}{c}{Duke}                 \\ \cline{1-5}
			Settings      & mAP                          & Rank-1                   & mAP  & Rank-1 \\ \hline
			Stage 1   & 45.0                          & 71.6                     & 41.4  & 62.9   \\
			Stage 2$^{*}$ & 26.6                         & 48.8                     & 7.2 & 16.7   \\
			Stage1 + Stage 2$^{*}$ & 55.1                         & 78.6                     & 35.8  & 54.2   \\
			Stage1 + Stage 2 + Eq. \eqref{jaccard}          & 69.1                        & 88.2                     & 57.0 & 75.7   \\ 
			Stage1 + Stage 2 + Eq. \eqref{eq:intersim}          & 72.1                         & 88.8                     & 59.1 & 76.9   \\ \hline
		\end{tabular}}
	\end{center}
	\caption{Ablation study on individual components of IICS. Stage 1 denotes intra-camera training stage. Stage 2 denotes inter-camera training stage. * denotes the CNN features similarity is used in stage 2.}
	\label{tab:components}
\end{table}


\vspace{-4mm}
\paragraph{The impact of AIBN.}
To test the validity of AIBN, we test it with different training settings.
The results are summarized in Table \ref{IBN}.
Replacing BNs in backbone with IN can improve the performance on DukeMTMC-ReID but decrease performance on Market1501, which shows that only applying IN can not bring stable performance enhancement.
AIBN can improve the performance on both dataset even though the mixture weight $\alpha$ of AIBN is fixed at 0.5 during training, which indicates that the combination of IN and BN brings more stable performance gains.
Optimizing mixture weight $\alpha$ can further improve the performance on Market1501 and DukeMTMC-ReID. It is clear that AIBN can improve the generalization ability of trained network on different domains and cameras.
IBN \cite{IBNNet} is another method of combining BN and IN to improve network generalization ability. The result shows that our AIBN substantially outperforms IBN.

\begin{table}
	\begin{center}
		\scalebox{0.9}{
		\begin{tabular}{c|cc|cc}
			\hline
			Dataset               & \multicolumn{2}{c|}{Market} & \multicolumn{2}{c}{Duke}                 \\ \cline{1-5}
			Settings              & mAP                         & Rank-1                   & mAP  & Rank-1 \\ \hline
			Backbone             & 67.1                        & 85.5                     & 51.4 & 71.3   \\
			+ IBN \cite{IBNNet}  & 59.8                        & 81.1                     & 35.4 & 56.3   \\
			+ IN		         & 59.6                        & 83.2                     & 53.0 & 72.7   \\
			+ AIBN (fixed)       & 70.7                        & 88.0                     & 56.9 & 75.2   \\
			+ AIBN               & 72.1                        & 88.8                     & 59.1 & 76.9   \\ \hline
		\end{tabular}}
	\end{center}
	\caption{Ablation study on AIBN. The ResNet-50 is used as backbone.}
	\label{IBN}
\end{table}

To further test the generalization of the network with AIBN, we train the network with labels on each dataset and test it directly on another dataset without fine-tuning.
The results are shown in Table \ref{Generalization}. On Market1501 and DukeMTMC-ReID, the AIBN improves the rank-1 accuracy by 5.6\% and 11.3\% for the direct transfer task, respectively.
It is clear that AIBN can improve the generalization of the network.
\begin{table}
	\begin{center}
		\setlength{\tabcolsep}{6pt}
		\scalebox{0.9}{
			\begin{tabular}{c|c|cc|cc}
				\hline
				\multirow{2}{*}{Training Set}  & Dataset     & \multicolumn{2}{c|}{Market } & \multicolumn{2}{c}{Duke}                 \\ \cline{2-6}
				                               & Settings      & mAP                              & Rank-1                            & mAP  & Rank-1 \\ \hline
				\multirow{2}{*}{Market}        & w/o AIBN  & 79.9                             & 92.0                              & 22.4 & 38.2   \\
				                               & w/ AIBN     & 80.0                             & 92.0                              & 29.5 & 49.5   \\ \hline
				\multirow{2}{*}{Duke}          & w/o AIBN  & 21.8                             & 48.9                              & 68.7 & 83.9   \\
				                               & w/ AIBN     & 27.2                             & 54.5                              & 69.2 & 84.8   \\ \hline
			\end{tabular}}
	\end{center}
	\vspace{-1mm}		
	\caption{Evaluation on the generalization ability of backbone with/without AIBN.}
	\label{Generalization}
\end{table}

\begin{figure}
	\centering
	\includegraphics[width=1\linewidth]{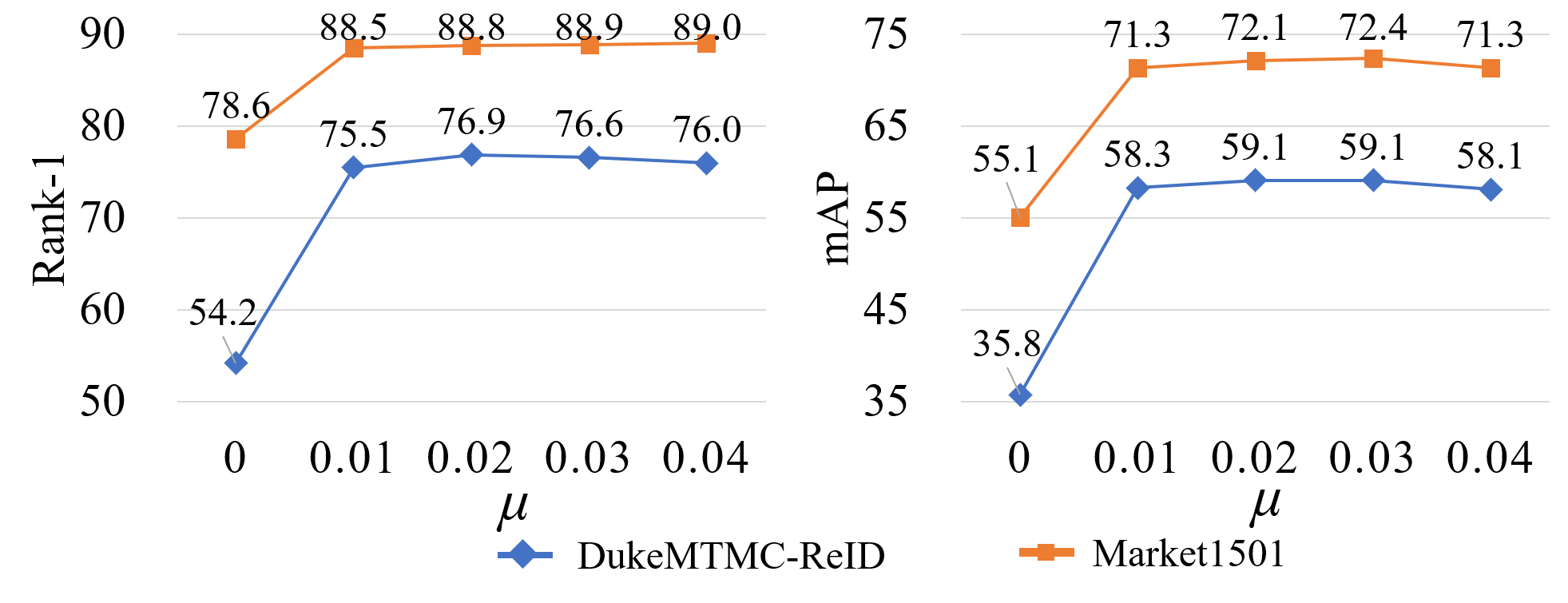}
	\vspace{-6mm}
	\caption{Evaluation of parameter $\mu$ in Eq. \eqref{eq:intersim}.}
	\label{margin}
\end{figure}

\vspace{-4mm}
\paragraph{Hyper-parameter Analysis.}
We investigate some important hyper-parameters in this section.
Fig. \ref{margin} shows effects of parameter $\mu$ in Eq. \eqref{eq:intersim}. We can see that, as $\mu$ increases from 0 to 0.02, rank-1 accuracy on Market1501 and DukeMTMC-ReID increases from 78.6\% and 55.1\% to 88.8\% and 72.1\%, respectively. This shows that larger u brings considerable performance gains. It is also clear that $\mu$ is easy to tune, i.e., $\mu > 0.01$ leads to similar performance on different datasets. 

We also conduct several experiments to verify the impacts of replacing BN with AIBN in different layers of the network and different weight sharing methods of $\alpha$ in Eq. \eqref{combine}. The results are shown in Table \ref{position}.

Replacing BNs in the deep layer of the network brings more substantial performance gains than replacing BNs in the shallow layer of the network.
It is clear that, replacing BNs with AIBN in Layer4 brings more significant performance gains than the replacements in Layer1 and Layer2. 
Since replacing BNs of Layer3 and Layer4 gives slightly better results, this setting is used in our experiments.
We evaluate three settings of weight sharing methods of $\alpha$: (a) Each AIBN has its own $\alpha$; (b) AIBNs in the same BottleNeck module share the same $\alpha$; (c) AIBNs in the same Layer of ResNet-50 share the same $\alpha$.
The results show that different weight sharing methods of $\alpha$ has limited impacts on the model performance.
\begin{table}
	\begin{center}
		\scalebox{0.9}{
		\begin{tabular}{c|c|cc|cc}
			\hline
			\multicolumn{2}{c|}{\multirow{2}{*}{Dataset}}   & \multicolumn{2}{c|}{Market} & \multicolumn{2}{c}{Duke}                 \\ \cline{3-6}
			\multicolumn{2}{l|}{}                         & mAP                         & Rank-1                   & mAP           & Rank-1          \\ \hline
			
		    \multirow{6}{*}{\rotatebox[origin=c]{90}{Position}} &Not Replace              & 67.1                        & 85.5                     & 51.4          & 71.3            \\
			& All                   & 72.1                        & 88.7                     & 58.7          & 77.1            \\
			& Layer 1+2             & 64.1                        & 86.2                     & 53.3          & 73.2            \\
			& Layer 4		        & 72.1                        & 88.5                     & 57.5          & 76.1            \\
			& Layer 2+3+4           & 71.3                        & 88.7                     & \textbf{59.8} & \textbf{77.4}   \\
			& Layer 3+4             & \textbf{72.1}               & \textbf{88.8}            & 59.1          & 76.9            \\ \hline
			
			\multirow{3}{*}{\rotatebox[origin=c]{90}{Sharing}}&Not sharing           & \textbf{72.4}               & \textbf{88.9}            & 58.4          & 76.2            \\
			& BottleNeck           & 72.1                        & 88.8                     & \textbf{59.1} & \textbf{76.9}   \\
			& Layer                 & 71.0                        & 88.3                     & 58.3          & 76.4            \\ \hline
		\end{tabular}}
	\end{center}
	\vspace{-1mm}
	\caption{Ablation study on inserted layer of AIBN and weight sharing methods of $\alpha$ in Eq. \eqref{combine}.}
	\label{position}
\end{table}

\vspace{-1mm}
\subsection{Comparison with State-of-the-art Methods}
We compare our method with recent unsupervised and transfer learning methods on Market1501 \cite{market1501}, DukeMTMC-ReID \cite{dukemtmc} and MSMT17 \cite{PTGan}.
Table \ref{sota} and Table \ref{msmt} summarize the comparison.
\begin{table*}
	\begin{center}
		\setlength{\tabcolsep}{6pt}
		\scalebox{0.83}{
		\begin{tabular}{c|c|ccccc|ccccc}
			\hline
			\multirow{2}{*}{Methods} & \multirow{2}{*}{Reference}           & \multicolumn{5}{c|}{Market1501 } & \multicolumn{5}{c}{DukeMTMC-ReID}                                                                                                                          \\ \cline{3-12}
			                         &           & Source                           & mAP                               & Rank-1        & Rank-5        & Rank-10       & Source & mAP           & Rank-1        & Rank-5        & Rank-10       \\ \hline
			PTGAN \cite{PTGan} &CVPR18                 & Duke                             & -                                 & 38.6          & -             & 66.1          & Market & -             & 27.4          & -             & 50.7          \\
			HHL \cite{HHL}  & ECCV18                    & Duke                             & 31.4                              & 62.2          & 78.8          & 84.0          & Market & 27.2          & 46.9          & 61.0          & 66.7          \\
			DG-Net++ \cite{zou2020joint} &ECCV20       & Duke                             & 61.7                              & 82.1          & 90.2          & 92.7          & Market & 63.8          & 78.9          & 87.8          & 90.4          \\ \hline
			TJ-AIDL \cite{wang2018transferable} & CVPR18 & Duke                             & 26.5                              & 58.2          & 74.8          & 81.8          & Market & 23.0          & 44.3          & 59.6          & 65.0          \\
			MMFA \cite{lin2018multi} & BMVC18           & Duke                             & 27.4                              & 56.7          & 75.0          & 81.8          & Market & 24.7          & 45.3          & 59.8          & 66.3          \\
			CSCL \cite{wu2019unsupervised} & ICCV19      & Duke                             & 35.6                              & 64.7          & 80.2          & 85.6          & Market & 30.5          & 51.5          & 66.7          & 71.7          \\ \hline
			MAR \cite{yu2019unsupervised} & CVPR19      & MSMT17                           & 40.0                              & 67.7          & 81.9          & -             & MSMT17 & 48.0          & 67.1          & 79.8          & -             \\
			AD-Cluster \cite{zhai2020ad} & CVPR20     & Duke                             & 68.3                              & 86.7          & 94.4          & 96.5          & Market & 54.1          & 72.6          & 82.5          & 85.5          \\
			NRMT \cite{zhao2020unsupervised} & ECCV20    & Duke                             & 71.7                              & 87.8          & 94.6          & 96.5          & Market & 62.2          & 77.8          & 86.9          & 89.5          \\
			MMT-500 \cite{MMT} &  ICLR20                & Duke                             & 71.2                              & 87.7          & 94.9          & 96.9          & Market & 63.1          & 76.8          & 88.0          & \textbf{92.2}          \\
			MEB-Net$^*$ \cite{zhai2020multiple} & ECCV20     & Duke                             & 71.9                & 87.5 & 95.2 & 96.8 & Market & 63.5 & 77.2          & 87.9          & 91.3 \\ \hline
			LOMO \cite{LOMO}& CVPR15                    & None                             & 8.0                               & 27.2          & 41.6          & 49.1          & None   & 4.8           & 12.3          & 21.3          & 26.6          \\
			BOW \cite{market1501}& ICCV15                      & None                             & 14.8                              & 35.8          & 52.4          & 60.3          & None   & 8.3           & 17.1          & 28.8          & 34.9          \\
			BUC \cite{lin2019bottom} & AAAI19            & None                             & 29.6                              & 61.9          & 73.5          & 78.2          & None   & 22.1          & 40.4          & 52.5          & 58.2          \\
			HCT \cite{zeng2020hierarchical} & CVPR20    & None                             & 56.4                              & 80.0          & 91.6          & 95.2          & None   & 50.7          & 69.6          & 83.4          & 87.4          \\
			MMCL \cite{multi-label} & CVPR20             & None                             & 45.5                              & 80.3          & 89.4          & 92.3          & None   & 40.2          & 65.2          & 75.9          & 80.0          \\
			JVTC+ \cite{JVTC} &ECCV20                  & None                             & 47.5                              & 79.5          & 89.2          & 91.9          & None   & 50.7          & 74.6          & 82.9          & 85.3          \\ \hline
			IICS$^\dag$ & This paper                  & None                             & 72.1                              & 88.8          & 95.3          & 96.9          & None   & 59.1          & 76.9          & 86.1          & 89.8          \\
			IICS$^\ddag$ & This paper                & None                             & \textbf{72.9}                      & \textbf{89.5}          & \textbf{95.2}          & \textbf{97.0}          & None   & \textbf{64.4}          & \textbf{80.0} & \textbf{89.0} & 91.6 \\ \hline
		\end{tabular}}
	\end{center}
	\vspace{-2mm}
	\caption{Performance comparison with recent methods on Market1501 and DukeMTMC-ReID. IICS denotes our method. \dag denotes using the cosine similarity to compute the CNN features similarity. \ddag denotes using the re-ranking similarity \cite{reranking} to replace the cosine similarly.
    * denotes the same backbone ResNet-50 is used in MEB-Net.}
	\label{sota}
\end{table*}

\begin{table}
	\begin{center}
		\scalebox{0.83}{
		\begin{tabular}{c|c|cccc}
			\hline
			\multirow{2}{*}{Methods}         & \multirow{2}{*}{Source} & \multicolumn{4}{c}{MSMT17}                             \\ \cline{3-6}
			                                 &                         & mAP                        & Rank-1 & Rank-5 & Rank-10 \\ \hline
			PTGAN \cite{PTGan}               & Market                  & 2.9                        & 10.2   & -      & 24.4    \\
			ECN \cite{ECN}                   & Market                  & 8.5                        & 25.3   & 36.3   & 42.1    \\
			SSG \cite{fu2019self}            & Market                  & 13.2                       & 31.6   & -      & 49.6    \\
			NRMT \cite{zhao2020unsupervised} & Market                  & 19.8                       & 43.7   & 56.5   & 62.2    \\
			DG-Net++ \cite{zou2020joint}     & Market                  & 22.1                       & 48.4   & 60.9   & 66.1    \\
			MMT-1500 \cite{MMT}              & Market                  & 22.9                       & 49.2   & 63.1   & 68.8    \\ \hline

			PTGAN \cite{PTGan}               & Duke                    & 3.3                        & 11.8   & -      & 27.4    \\
			ECN \cite{ECN}                   & Duke                    & 10.2                       & 30.2   & 41.5   & 46.8    \\
			SSG \cite{fu2019self}            & Duke                    & 13.3                       & 32.2   & -      & 51.2    \\
			NRMT \cite{zhao2020unsupervised} & Duke                    & 20.6                       & 45.2   & 57.8   & 63.3    \\
			DG-Net++ \cite{zou2020joint}     & Duke                    & 22.1                       & 48.8   & 60.9   & 65.9    \\
			MMT-1500 \cite{MMT}              & Duke                    & 23.3                       & 50.1   & 63.9   & 69.8    \\ \hline

			MMCL \cite{multi-label}          & None                    & 11.2                       & 35.4   & 44.8   & 49.8    \\
			JVTC+ \cite{JVTC}                & None                    & 17.3                       & 43.1   & 53.8   & 59.4    \\ \hline
			IICS$^\dag$                & None                    & 18.6                       & 45.7   & 57.7   & 62.8    \\
			IICS$^\ddag$               & None                    & \textbf{26.9}              & \textbf{56.4}   & \textbf{68.8}   & \textbf{73.4}    \\ \hline
		\end{tabular}}
	\end{center}
	\vspace{-2mm}
	\caption{Performance comparison with recent methods on MSMT17 \cite{PTGan}. IICS denotes our method. \dag denotes using the cosine similarity to compute the CNN feature similarity. \ddag denotes using the re-ranking similarity \cite{reranking} to replace the cosine similarity.}
	\label{msmt}
\end{table}


We first compare our method with methods trained with only unlabeled data. Compared methods include hand-craft features based methods, and deep learning based methods. It can be seen from Table \ref{sota} that compared with other deep learning based methods, our method surpasses these methods by a large margin. This significant improvement is mainly thanks to the more reliable similarity between samples used in clustering.

We also compare with the unsupervised domain adaptation methods, including GAN based methods (PTGAN \cite{PTGan}, \etc), Distribution alignment based methods (TJ-AIDL \cite{wang2018transferable}, \etc), and Pseudo-labels based methods (MAR \cite{yu2019unsupervised}, \etc).
Pseudo-labels based methods perform better than other types of methods in most cases.
Many transfer learning methods use extra labeled source domain data for training. Our method still outperforms them using only unlabeled data for training.  
The performance of our method can be further improved by using the re-ranking similarity \cite{reranking} instead of the cosine similarity. Note that, re-ranking similarity is a commonly used similarity in unsupervised ReID \cite{fu2019self, JVTC} and is only used during training. Therefore, it only increases the training time and has no effect on the network inference time and online ReID time.

To further verify the effectiveness of our algorithm, we conduct experiments on a larger and more challenging dataset MSMT17.
Our method outperforms existing methods under both unsupervised and unsupervised transfer settings by a large margin. We achieve the rank-1 accuracy of 56.4\%, about 11\% higher than the recent NRMT \cite{zhao2020unsupervised}, which adopts extra DukeMTMC-ReID for training. Those above experiments clearly demonstrate the superior performance of the proposed method.
\vspace{-4mm}
\paragraph{Discussion} Our method uses pre-defined clustering numbers in both stages, thus the clustering number is a critical parameter for pseudo label generation.
The clustering number can also be adaptively determined by setting a similarity threshold. Generalizable strategies for determining the clustering number for different datasets will be studied in our future work.

\vspace{-2mm}
\section{Conclusion}
This paper proposes a intra-inter camera similarity method for unsupervised person ReID which iteratively optimizes Intra-Inter Camera similarity through generating intra- and inter-camera pseudo-labels.
The intra-camera training stage is proposed to train a multi-branch CNN using generated intra-camera pseudo-labels.
Based on the classification score produced by each classifier trained at intra-camera training stage, a more robust inter-camera similarity can be calculated. Then the network can be trained with the pseudo-label generated by performing clustering across cameras with this inter-camera similarity.
Moreover, AIBN is introduced to boost the generalization ability of the network.
Extensive experimental results demonstrate the effectiveness of the proposed method in unsupervised person ReID.

\vspace{-4mm}
\paragraph{Acknowledgement} This work is supported in part by Peng Cheng Laboratory, in part by Natural Science Foundation of China under Grant No. U20B2052, 61936011, 61620106009, in part by The National Key Research and Development Program of China under Grant No. 2018YFE0118400, in part by Beijing Natural Science Foundation under Grant No. JQ18012.
\cleardoublepage

{\small
	\bibliographystyle{ieee_fullname}
	\bibliography{egbibb}

\begin{thebibliography}{10}\itemsep=-1pt

\bibitem{chang2019domain}
Woong-Gi Chang, Tackgeun You, Seonguk Seo, Suha Kwak, and Bohyung Han.
\newblock Domain-specific batch normalization for unsupervised domain
  adaptation.
\newblock In {\em CVPR}, 2019.

\bibitem{imagenet}
Jia Deng, Wei Dong, Richard Socher, Li-Jia Li, Kai Li, and Li Fei-Fei.
\newblock Imagenet: A large-scale hierarchical image database.
\newblock In {\em CVPR}, 2009.

\bibitem{deng2018image}
Weijian Deng, Liang Zheng, Qixiang Ye, Guoliang Kang, Yi Yang, and Jianbin
  Jiao.
\newblock Image-image domain adaptation with preserved self-similarity and
  domain-dissimilarity for person re-identification.
\newblock In {\em CVPR}, 2018.

\bibitem{dou2019domain}
Qi Dou, Daniel~Coelho de Castro, Konstantinos Kamnitsas, and Ben Glocker.
\newblock Domain generalization via model-agnostic learning of semantic
  features.
\newblock In {\em NeurIPS}, 2019.

\bibitem{fan2018unsupervised}
Hehe Fan, Liang Zheng, Chenggang Yan, and Yi Yang.
\newblock Unsupervised person re-identification: Clustering and fine-tuning.
\newblock {\em ACM Transactions on Multimedia Computing, Communications, and
  Applications (TOMM)}, 14(4):1--18, 2018.

\bibitem{fu2019self}
Yang Fu, Yunchao Wei, Guanshuo Wang, Yuqian Zhou, Honghui Shi, and Thomas~S
  Huang.
\newblock Self-similarity grouping: A simple unsupervised cross domain
  adaptation approach for person re-identification.
\newblock In {\em ICCV}, 2019.

\bibitem{MMT}
Yixiao Ge, Dapeng Chen, and Hongsheng Li.
\newblock Mutual mean-teaching: Pseudo label refinery for unsupervised domain
  adaptation on person re-identification.
\newblock In {\em ICLR}, 2020.

\bibitem{MMD}
Arthur Gretton, Kenji Fukumizu, Zaid Harchaoui, and Bharath~K Sriperumbudur.
\newblock A fast, consistent kernel two-sample test.
\newblock In {\em NeurIPS}, 2009.

\bibitem{resnet}
Kaiming He, Xiangyu Zhang, Shaoqing Ren, and Jian Sun.
\newblock Deep residual learning for image recognition.
\newblock In {\em CVPR}, 2016.

\bibitem{triplet}
Alexander Hermans, Lucas Beyer, and Bastian Leibe.
\newblock In defense of the triplet loss for person re-identification.
\newblock {\em arXiv preprint arXiv:1703.07737}, 2017.

\bibitem{BN}
Sergey Ioffe and Christian Szegedy.
\newblock Batch normalization: Accelerating deep network training by reducing
  internal covariate shift.
\newblock In {\em ICML}, 2015.

\bibitem{jin2020global}
Xin Jin, Cuiling Lan, Wenjun Zeng, and Zhibo Chen.
\newblock Global distance-distributions separation for unsupervised person
  re-identification.
\newblock In {\em ECCV}, 2020.

\bibitem{JVTC}
Jianing Li and Shiliang Zhang.
\newblock Joint visual and temporal consistency for unsupervised domain
  adaptive person re-identification.
\newblock In {\em ECCV}, 2020.

\bibitem{LOMO}
Shengcai Liao, Yang Hu, Xiangyu Zhu, and Stan~Z Li.
\newblock Person re-identification by local maximal occurrence representation
  and metric learning.
\newblock In {\em CVPR}, 2015.

\bibitem{lin2018multi}
Shan Lin, Haoliang Li, Chang-Tsun Li, and Alex~Chichung Kot.
\newblock Multi-task mid-level feature alignment network for unsupervised
  cross-dataset person re-identification.
\newblock In {\em BMVC}, 2018.

\bibitem{lin2019bottom}
Yutian Lin, Xuanyi Dong, Liang Zheng, Yan Yan, and Yi Yang.
\newblock A bottom-up clustering approach to unsupervised person
  re-identification.
\newblock In {\em AAAI}, 2019.

\bibitem{liu2020domain}
Xiaobin Liu and Shiliang Zhang.
\newblock Domain adaptive person re-identification via coupling optimization.
\newblock In {\em Proceedings of the 28th ACM International Conference on
  Multimedia}, pages 547--555, 2020.

\bibitem{liu2019self}
Xiaobin Liu, Shiliang Zhang, and Ming Yang.
\newblock Self-guided hash coding for large-scale person re-identification.
\newblock In {\em MIPR}, pages 246--251. IEEE, 2019.

\bibitem{LuoSB}
H. {Luo}, W. {Jiang}, Y. {Gu}, F. {Liu}, X. {Liao}, S. {Lai}, and J. {Gu}.
\newblock A strong baseline and batch normalization neck for deep person
  re-identification.
\newblock {\em IEEE Transactions on Multimedia}, 22(10):2597--2609, 2020.

\bibitem{t-SNE}
Laurens van~der Maaten and Geoffrey Hinton.
\newblock Visualizing data using t-sne.
\newblock {\em Journal of machine learning research}, 9(Nov):2579--2605, 2008.

\bibitem{IBNNet}
Xingang Pan, Ping Luo, Jianping Shi, and Xiaoou Tang.
\newblock Two at once: Enhancing learning and generalization capacities via
  ibn-net.
\newblock In {\em ECCV}, 2018.

\bibitem{scikit-learn}
F. Pedregosa, G. Varoquaux, A. Gramfort, V. Michel, B. Thirion, O. Grisel, M.
  Blondel, P. Prettenhofer, R. Weiss, V. Dubourg, J. Vanderplas, A. Passos, D.
  Cournapeau, M. Brucher, M. Perrot, and E. Duchesnay.
\newblock Scikit-learn: Machine learning in {P}ython.
\newblock {\em Journal of Machine Learning Research}, 12:2825--2830, 2011.

\bibitem{dukemtmc}
Ergys Ristani, Francesco Solera, Roger Zou, Rita Cucchiara, and Carlo Tomasi.
\newblock Performance measures and a data set for multi-target, multi-camera
  tracking.
\newblock In {\em ECCV}, 2016.

\bibitem{coral}
Baochen Sun and Kate Saenko.
\newblock Deep coral: Correlation alignment for deep domain adaptation.
\newblock In {\em ECCV}, 2016.

\bibitem{PCB}
Yifan Sun, Liang Zheng, Yi Yang, Qi Tian, and Shengjin Wang.
\newblock Beyond part models: Person retrieval with refined part pooling (and a
  strong convolutional baseline).
\newblock In {\em ECCV}, 2018.

\bibitem{tzeng2015simultaneous}
Eric Tzeng, Judy Hoffman, Trevor Darrell, and Kate Saenko.
\newblock Simultaneous deep transfer across domains and tasks.
\newblock In {\em ICCV}, 2015.

\bibitem{IN}
Dmitry Ulyanov, Andrea Vedaldi, and Victor Lempitsky.
\newblock Improved texture networks: Maximizing quality and diversity in
  feed-forward stylization and texture synthesis.
\newblock In {\em CVPR}, 2017.

\bibitem{multi-label}
Dongkai Wang and Shiliang Zhang.
\newblock Unsupervised person re-identification via multi-label classification.
\newblock In {\em CVPR}, 2020.

\bibitem{wang2018transferable}
Jingya Wang, Xiatian Zhu, Shaogang Gong, and Wei Li.
\newblock Transferable joint attribute-identity deep learning for unsupervised
  person re-identification.
\newblock In {\em CVPR}, 2018.

\bibitem{demo}
Longhui Wei, Xiaobin Liu, Jianing Li, and Shiliang Zhang.
\newblock Vp-reid: Vehicle and person re-identification system.
\newblock In {\em ICMR}, ICMR '18, page 501–504, New York, NY, USA, 2018.
  Association for Computing Machinery.

\bibitem{PTGan}
Longhui Wei, Shiliang Zhang, Wen Gao, and Qi Tian.
\newblock Person transfer gan to bridge domain gap for person
  re-identification.
\newblock In {\em CVPR}, 2018.

\bibitem{wu2019unsupervised}
Ancong Wu, Wei-Shi Zheng, and Jian-Huang Lai.
\newblock Unsupervised person re-identification by camera-aware similarity
  consistency learning.
\newblock In {\em ICCV}, 2019.

\bibitem{yu2019unsupervised}
Hong-Xing Yu, Wei-Shi Zheng, Ancong Wu, Xiaowei Guo, Shaogang Gong, and
  Jian-Huang Lai.
\newblock Unsupervised person re-identification by soft multilabel learning.
\newblock In {\em CVPR}, 2019.

\bibitem{zeng2020hierarchical}
Kaiwei Zeng, Munan Ning, Yaohua Wang, and Yang Guo.
\newblock Hierarchical clustering with hard-batch triplet loss for person
  re-identification.
\newblock In {\em CVPR}, 2020.

\bibitem{zhai2020ad}
Yunpeng Zhai, Shijian Lu, Qixiang Ye, Xuebo Shan, Jie Chen, Rongrong Ji, and
  Yonghong Tian.
\newblock Ad-cluster: Augmented discriminative clustering for domain adaptive
  person re-identification.
\newblock In {\em CVPR}, 2020.

\bibitem{zhai2020multiple}
Yunpeng Zhai, Qixiang Ye, Shijian Lu, Mengxi Jia, Rongrong Ji, and Yonghong
  Tian.
\newblock Multiple expert brainstorming for domain adaptive person
  re-identification.
\newblock In {\em ECCV}, 2020.

\bibitem{zhang2019self}
Xinyu Zhang, Jiewei Cao, Chunhua Shen, and Mingyu You.
\newblock Self-training with progressive augmentation for unsupervised
  cross-domain person re-identification.
\newblock In {\em ICCV}, 2019.

\bibitem{zhang2017alignedreid}
Xuan Zhang, Hao Luo, Xing Fan, Weilai Xiang, Yixiao Sun, Qiqi Xiao, Wei Jiang,
  Chi Zhang, and Jian Sun.
\newblock Alignedreid: Surpassing human-level performance in person
  re-identification.
\newblock {\em arXiv preprint arXiv:1711.08184}, 2017.

\bibitem{mutual}
Ying Zhang, Tao Xiang, Timothy~M Hospedales, and Huchuan Lu.
\newblock Deep mutual learning.
\newblock In {\em CVPR}, 2018.

\bibitem{zhao2020unsupervised}
Fang Zhao, Shengcai Liao, Guo-Sen Xie, Jian Zhao, Kaihao Zhang, and Ling Shao.
\newblock Unsupervised domain adaptation with noise resistible mutual-training
  for person re-identification.
\newblock In {\em ECCV}, 2020.

\bibitem{market1501}
Liang Zheng, Liyue Shen, Lu Tian, Shengjin Wang, Jingdong Wang, and Qi Tian.
\newblock Scalable person re-identification: A benchmark.
\newblock In {\em ICCV}, 2015.

\bibitem{reranking}
Zhun Zhong, Liang Zheng, Donglin Cao, and Shaozi Li.
\newblock Re-ranking person re-identification with k-reciprocal encoding.
\newblock In {\em CVPR}, 2017.

\bibitem{HHL}
Zhun Zhong, Liang Zheng, Shaozi Li, and Yi Yang.
\newblock Generalizing a person retrieval model hetero-and homogeneously.
\newblock In {\em ECCV}, 2018.

\bibitem{ECN}
Zhun Zhong, Liang Zheng, Zhiming Luo, Shaozi Li, and Yi Yang.
\newblock Invariance matters: Exemplar memory for domain adaptive person
  re-identification.
\newblock In {\em CVPR}, 2019.

\bibitem{zhong2018camera}
Zhun Zhong, Liang Zheng, Zhedong Zheng, Shaozi Li, and Yi Yang.
\newblock Camera style adaptation for person re-identification.
\newblock In {\em CVPR}, 2018.

\bibitem{CycleGAN2017}
Jun-Yan Zhu, Taesung Park, Phillip Isola, and Alexei~A Efros.
\newblock Unpaired image-to-image translation using cycle-consistent
  adversarial networkss.
\newblock In {\em ICCV}, 2017.

\bibitem{zhu2021intracamera}
Xiangping Zhu, Xiatian Zhu, Minxian Li, Pietro Morerio, Vittorio Murino, and
  Shaogang Gong.
\newblock Intra-camera supervised person re-identification.
\newblock {\em arXiv preprint arXiv:2002.05046}, 2021.

\bibitem{zhuangrethinking}
Zijie Zhuang, Longhui Wei, Lingxi Xie, Tianyu Zhang, Hengheng Zhang, Haozhe Wu,
  Haizhou Ai, and Qi Tian.
\newblock Rethinking the distribution gap of person re-identification with
  camera-based batch normalization.
\newblock In {\em ECCV}, 2020.

\bibitem{zou2020joint}
Yang Zou, Xiaodong Yang, Zhiding Yu, BVK Kumar, and Jan Kautz.
\newblock Joint disentangling and adaptation for cross-domain person
  re-identification.
\newblock In {\em ECCV}, 2020.

\end{thebibliography}
}

\end{document}